\documentclass[letterpaper]{article} 
\usepackage{aaai2026}  
\usepackage{times}  
\usepackage{helvet}  
\usepackage{courier}  
\usepackage[hyphens]{url}  
\usepackage{graphicx} 
\usepackage{subcaption}
\usepackage{booktabs}
\usepackage{natbib}  
\usepackage{caption} 
\usepackage{algorithm}
\usepackage{algorithmic}
\usepackage{mathtools}
\usepackage{amsmath}
\usepackage{multirow, makecell}
\usepackage{graphicx, bm}
\usepackage{amsthm}
\newcommand{\vw}{\bm{w}}
\newcommand{\vx}{\bm{x}}
\newcommand{\vr}{\bm{r}}
\newcommand{\vo}{\bm{o}}
\newcommand{\mW}{\bm{W}}
\newcommand{\mX}{\bm{X}}
\newcommand{\sR}{\mathcal{R}}
\newcommand{\gD}{\mathcal{D}}
\newcommand{\gN}{\mathcal{N}}
\newcommand{\sN}{\mathcal{N}}
\newcommand{\sP}{\mathcal{P}}
\newcommand{\E}{\mathcal{E}}
\newcommand{\gB}{\mathcal{B}}

\usepackage{newfloat}
\usepackage{listings}
\DeclareCaptionStyle{ruled}{labelfont=normalfont,labelsep=colon,strut=off} 
\floatstyle{ruled}
\newfloat{listing}{tb}{lst}{}
\floatname{listing}{Listing}
\makeatletter
\def\copyright@on{F}
\def\showauthors@on{T}
\makeatother
\title{Expert-Token Resonance MoE: Bidirectional Routing with Efficiency Affinity-Driven Active Selection}
\author{
	Jing Li\equalcontrib,
	Zhijie Sun\equalcontrib,
	Dachao Lin\equalcontrib,
	Xuan He,
	Binfan Zheng,
	Yi Lin,
	Rongqian Zhao,
	Xin Chen
}
\affiliations{
    \textsuperscript{\rm 1}Huawei Technologies Co., Ltd\\

}

\usepackage{bibentry}

\begin{document}

\maketitle

\begin{abstract}
	Mixture-of-Experts (MoE) architectures enable efficient scaling of large language models by activating only a subset of parameters per input. However, existing MoE models suffer from two critical limitations: (1) inefficient token-to-expert routing that causes excessive communication overhead, and (2) expert homogenization that leads to redundant computations. Current approaches address these challenges separately, failing to achieve simultaneous improvements in both training efficiency and model performance. We present Expert-Token Resonance (ETR), a theoretically-grounded bidirectional routing mechanism that fundamentally reimagines expert-token interactions in MoE architectures. Our key insight is that optimal routing requires adaptive coordination between token-choice routing (TCR) during early training phases and expert-choice routing (ECR) in later stages. We prove that this dynamic approach maximizes training success rate—the probability of correct token-expert assignments—while reducing the expert capacity lower bound by up to 40\%. ETR incorporates three technical innovations: (1) an affinity-based routing architecture using Grouped Average Pooling (GrAP) that reduces computational complexity from O(d²) to O(d²/D) while maintaining orthogonality to prevent expert homogenization; (2) a bidirectional selection mechanism that enables both tokens and experts to actively participate in the routing process based on cosine similarity scores; and (3) an adaptive capacity strategy that dynamically adjusts expert bounds based on training progress, eliminating communication bubbles in All-to-All operations. Extensive experiments on Ascend NPU clusters demonstrate that ETR achieves 5.4\%-46.6\% improvements in end-to-end training efficiency compared to baseline MoE implementations, with 9.7\%-14.5\% performance gains across GDAD, GPQA, HumanEval, and TeleQnA benchmarks. These results establish ETR as the efficient MoE routing approach to deliver substantial improvements in both computational efficiency and model quality, enabling practical deployment of larger sparse models previously constrained by communication bottlenecks.
\end{abstract}

\section{Introduction}

Large language models (LLMs) have demonstrated remarkable capabilities in understanding complex semantic relationships and generating coherent text \citep{zhao2023survey}. However, as model parameters scale to billions or even trillions, the computational and communication costs grow prohibitively \citep{jiang2024megascale}. The Mixture-of-Experts (MoE) architecture offers a promising solution by activating only a subset of model parameters for each input, enabling efficient model scaling without proportional increases in computational demands \citep{lepikhin2020gshard, fedus2022switch}. Recent MoE-based LLMs, such as DeepSeek-V3 \citep{liu2024deepseek} and Mixtral \citep{jiang2024mixtral}, have achieved state-of-the-art performance across various benchmarks.

Despite these successes, MoE models face two fundamental challenges that limit their practical deployment and effectiveness. First, \textbf{the routing efficiency problem}: current routing mechanisms suffer from suboptimal token-to-expert assignments, leading to significant computational waste through excessive padding in All-to-All communications and underutilized expert capacity. Second, \textbf{the expert homogenization problem}: existing routing strategies often fail to maintain expert specialization, resulting in redundant computations where multiple experts learn similar representations. These challenges are particularly pronounced during early training phases when routing networks lack sophisticated assignment capabilities.

Previous approaches have attempted to address these issues separately. Load balancing techniques \citep{zhou2022mixture, dai2022stablemoe} focus on distributing tokens evenly across experts but often sacrifice routing quality. Expert specialization methods \citep{xie2024mode, li2024locmoe} aim to maintain diversity but typically incur additional computational overhead. Critically, no existing work has demonstrated how to simultaneously improve both training efficiency and model performance through a unified routing mechanism.

This paper introduces Expert-Token Resonance (ETR), a novel bidirectional routing strategy that fundamentally rethinks how tokens and experts interact in MoE architectures. Our key insight is that optimal routing requires not only tokens selecting appropriate experts (token-choice routing, TCR) but also experts actively selecting their most relevant tokens (expert-choice routing, ECR). We theoretically prove that TCR achieves higher training success rates during early training when routing networks need refinement, while ECR excels in later stages when expert specialization becomes critical. By adaptively combining both paradigms, ETR achieves what previous methods could not: simultaneous improvements in training efficiency and model performance.

Specifically, our contributions are threefold:

\begin{enumerate}
	\item \textbf{A theoretically-grounded bidirectional routing mechanism} that dynamically balances token-choice and expert-choice routing based on training progress. We prove that this approach maximizes the training success rate—the probability of correct token-expert assignments—throughout the entire training process while reducing the required expert capacity lower bound by up to 40\% compared to conventional methods.
	
	\item \textbf{An efficient affinity-based routing architecture} leveraging Grouped Average Pooling (GrAP) layers that reduces computational complexity from $O(d^2)$ to $O(d^2/D)$ compared to traditional MLP routers, where $d$ is the hidden dimension and $D$ is the grouping factor. The orthogonality properties of GrAP naturally prevent expert homogenization while enabling precise affinity computations through cosine similarity.
	
	\item \textbf{Comprehensive empirical validation} on Ascend NPU clusters demonstrating that ETR improves end-to-end training efficiency by 5.4\%-46.6\% compared to baseline MoE implementations and 2.9\%-13.3\% compared to state-of-the-art LocMoE, while simultaneously enhancing downstream task performance by 9.7\%-14.5\% and 1.7\%-4.1\%, respectively.
\end{enumerate}

The significance of this work extends beyond incremental improvements. By addressing the fundamental trade-off between routing quality and computational efficiency, ETR enables practical deployment of larger MoE models that were previously constrained by communication bottlenecks. Our theoretical analysis provides new insights into the dynamics of expert-token interactions, opening avenues for future research in sparse model architectures.

\section{Related Work}

\subsection{Routing Mechanisms}
Routing design fundamentally determines MoE performance. Token-choice routing (TCR) \citep{shazeer2017outrageouslylargeneuralnetworksthe} allows tokens to select experts but suffers from load imbalance. Expert-choice routing (ECR) \citep{zhou2022mixtureofexpertswithexpertchoicerouting} ensures balanced loads by having experts select tokens, risking the loss of critical tokens. Recent dynamic routing strategies \citep{huang2024hardertasksneedmoreexperts} adapt expert allocation to input complexity but remain unidirectional—either tokens choose experts or vice versa. This fundamental limitation prevents simultaneous optimization of routing quality and load balance. 

\subsection{Expert Specialization}
Expert homogenization critically limits MoE performance. Prior work addressed this through fine-grained expert segmentation \citep{dai2024deepseekmoetowardsultimateexpertspecialization}, competitive mechanisms \citep{pham2024competesmoeeffectivetrainingofsparse}, and theoretical analyses linking data clustering to specialization \citep{chen2022towardsunderstandingmixtureofexperts}. However, these approaches require auxiliary losses or complex training procedures. 

\subsection{System Optimization}
System-level MoE optimizations focus on hardware utilization through block-sparse operations \citep{gale2022megablocksefficientsparsetrainingwith} and communication reduction via expert sharding \citep{balmau2025acceleratingmoemodelinferencewith} or sequence migration \citep{chen2024communicationefficientsparselyactivatedmodeltrainingvia}. While these approaches improve specific bottlenecks, they overlook communication bubbles in All-to-All operations—a critical inefficiency.

\subsection{Positioning of This Work}
Recent surveys \citep{cai2024asurveyonmixtureof} reveal that existing MoE methods typically prioritize either routing quality or computational efficiency. Our ETR framework addresses previously overlooked communication inefficiencies and training bottleneck, suggesting potential for more efficient MoE deployment at scale.

\section{Method}\label{method}


In this section, we present the efficient routing mechanism, and our adaptive bidirectional selection mechanism is detailed. Then, for traditional drop-and-pad strategies, a dynamic token distribution analysis module that optimizes the lower bounds of expert capacity are displayed. Moreover, we also describe the loss for expert load balancing.

\begin{figure}[!t]
	\centering
	\includegraphics[width=0.48\textwidth]{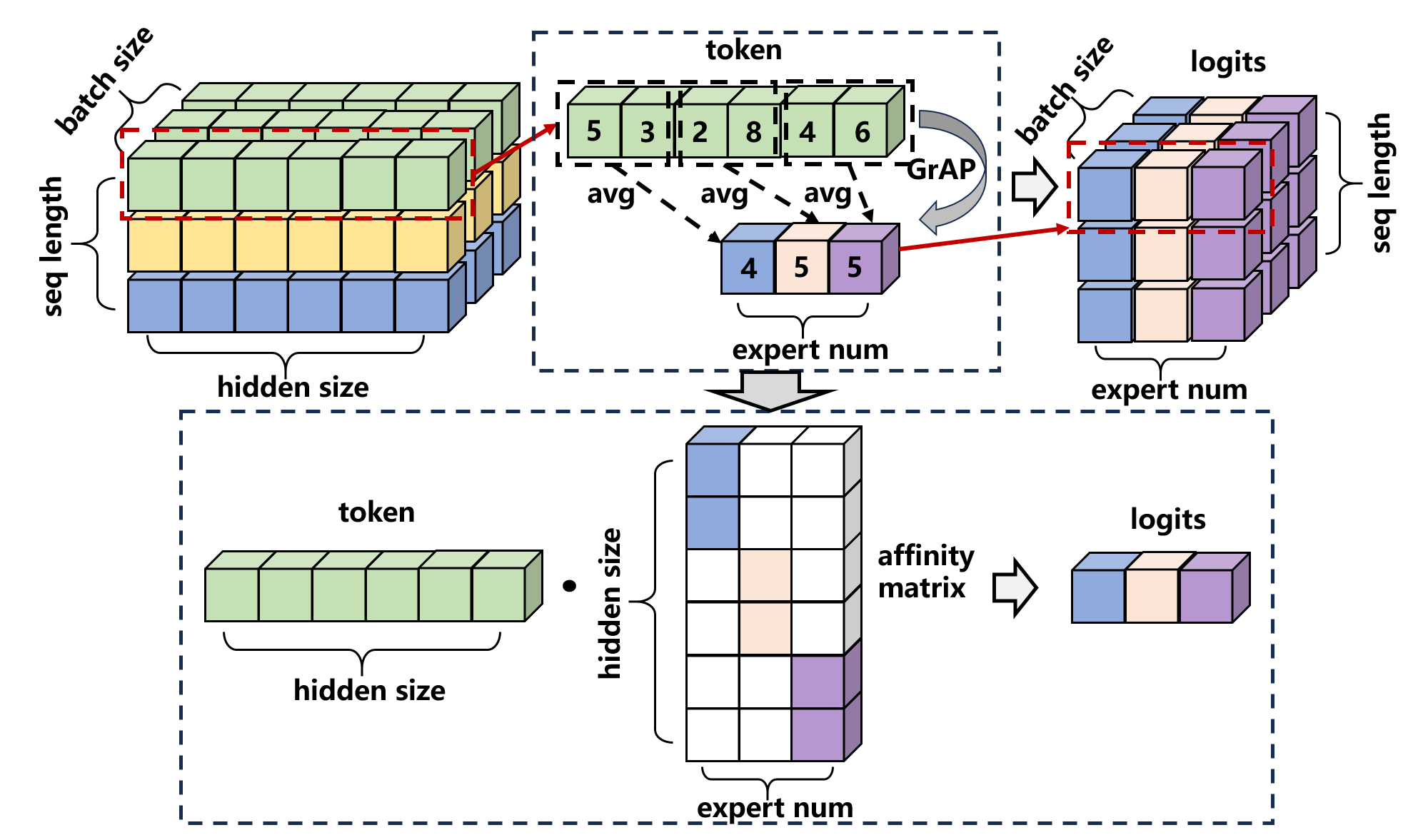}
	\caption{The illustrative diagram of GrAP.}
	\label{grap}
\end{figure}

\subsection{Model Architecture.}

\paragraph{Backbone.}

The MoE architecture, based on the Transformer framework, efficiently scales up model size with low computational overhead, benefiting from two primary structures: a sparse gating network for routing tokens and expert networks for processing specific token categories.

We consider the supervised classification for brevity where the training samples are $\{ (\vx^{(i)}, y_i) \}_{i=1}^N \sim \gD$. Each training sample $\vx^\top = (\vx_1^\top, \dots, \vx_s^\top) \in \sR^{sd}$ has $s$ tokens with token feature $\vx_i \in \sR^d, \forall i \in [s]$, and label $y \in \sN^+$.
The objective is to learn the map of $\vx$ to the corresponding $y$. The general MoE structure are formulated as 
\begin{equation}
	\small	\text{MoE}(\vx) = \sum_{t=1}^s \sum_{i=1}^n G_i(\vx_t) \cdot E_ i(\vx_t),
\end{equation}
where $n$ is the number of experts, $G(\vx_t) \colon \sR^{d} \to \sR^n$ is the gating weight vector of experts which maps the tokens of $\vx_t$ into the coresponding experts with weights, e.g., $G_i(\vx) = \text{Softmax}(\mW\vx+\bm{\epsilon})$ where the softmax is applied to each row, and $E_i(\vx_t) \colon \sR^d \to \sR$ is the $i$-th expert network, see \cite{liu2024routers} for current different router methods. Generally, $n \ll s$, which saves much computation compared to the dense structure.

\paragraph{Cost-Efficient Sparse Expert-Token Affinity.}

$\mW_{\text{aff}}$ denotes the \textbf{expert-token affinity matrix}. After processing through the GrAP routing layer, tokens generate a diagonal sparse matrix as shown. Compared to the dense matrix produced by traditional routing layers, this reduces the parameter count to 1/D of the original, significantly decreasing the computational overhead of the expert routing layer. 

\begin{figure}[H]
	\centering
	\includegraphics[width=0.48\textwidth]{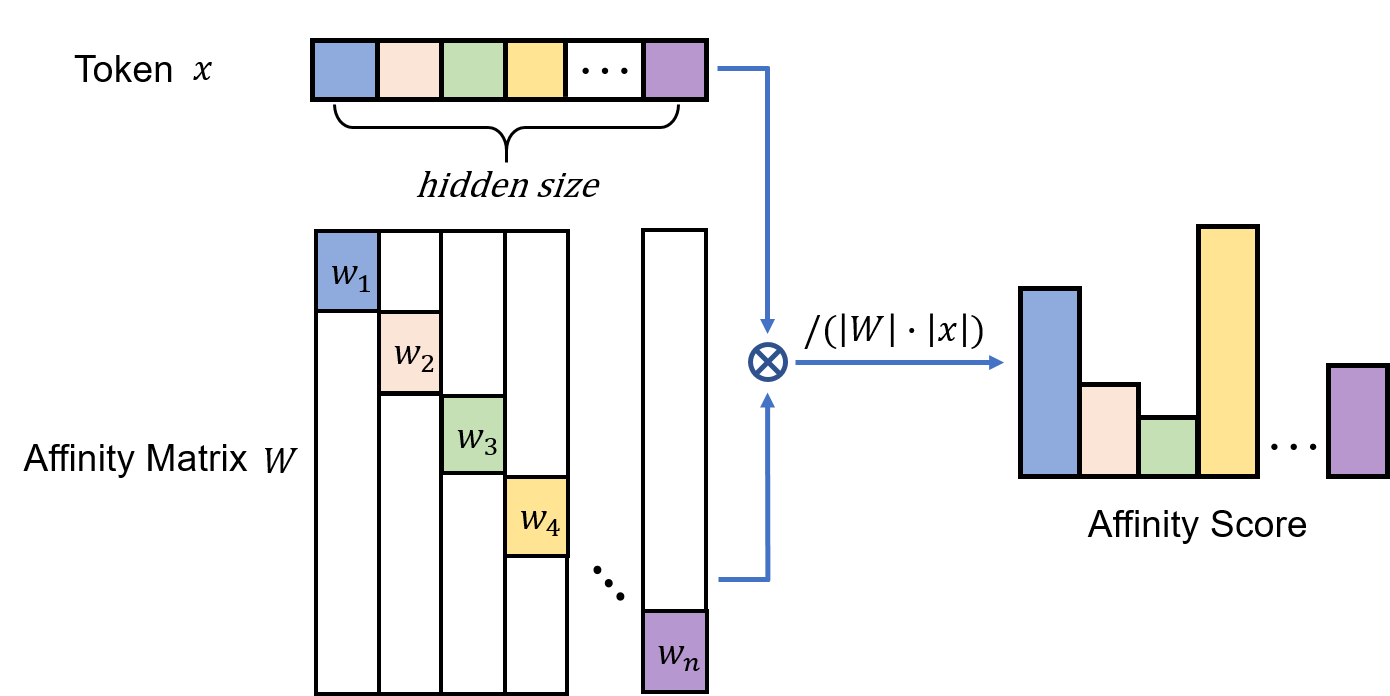}
	\caption{The illustration of affinity score.}
	\label{affmetric}
\end{figure}

As shown in Figure \ref{grap}, GrAP performs average pooling with group segmented by the number of experts. With GrAP as the layer of feature extraction, the formulation of $\mW_{\text{aff}}$ is as followed:
\begin{equation}
	\small \mW_{\text{aff}}  = \begin{pmatrix}
		\vw_1 & 0 & \cdots & 0 \\
		0 & \vw_2 & \cdots & 0 \\
		\vdots & \vdots & \ddots & \vdots \\
		0 & 0 & \cdots & \vw_n \\
	\end{pmatrix}
\end{equation}
\begin{equation}
	\small \vw_i = \frac{n}{d} \cdot \mathbf{1}\left\{\frac{i \cdot d}{n} \leq j < \left(i+1\right)\frac{d}{n}\right\} \quad 0 \leq j < d
\end{equation}
The expert-token affinity matrix is employed as the gating weight to calculate the affinity score between each expert and token. We define the affinity score of $t$-th token and $i$-th expert as the cosine similarity between vectors $\vx_t$ and  $\vw_i$:
\begin{equation}
	\small \delta_{ti} = \cos\left(\vx_t, \vw_i\right) := \vx_t^\top\vw_i / (\|\vx_t\|\cdot \|\vw_i\|)
\end{equation}
The affinity score intuitively reflect how closely the two inputs are associated. From a perspective of semantic, the affinity scores derived from affinity metrics consisting of orthogonal vectors represent the degree of association between each token and various experts, as shown in Figure \ref{affmetric}. Therefore, we leverage the affinity score as the principle of our affinity-driven active selection routing mechanism. 

\begin{figure}[t!]
	\centering
	\includegraphics[width=0.48\textwidth]{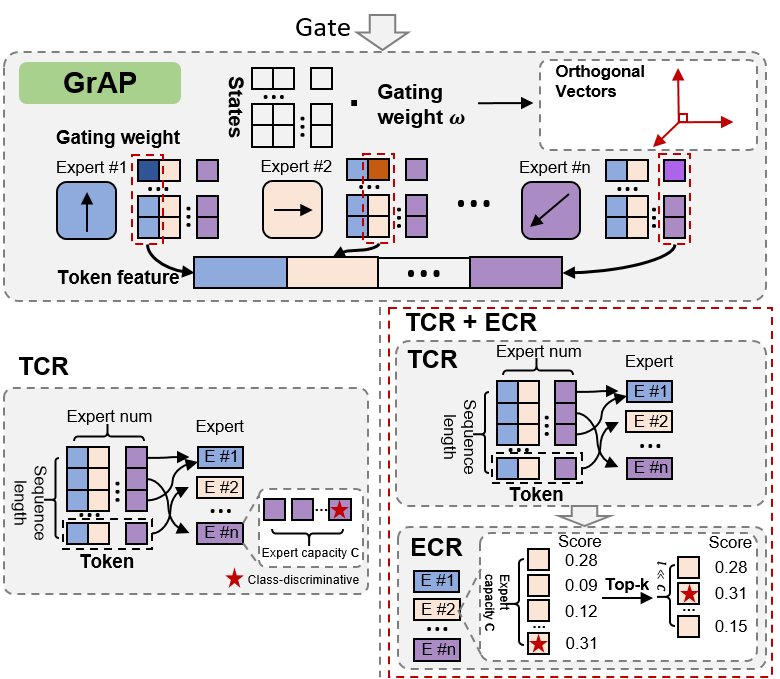}
	\caption{The architecture of the gate network along with the hybrid TCR + ECR router.}
	\label{arc}
\end{figure}

\paragraph{Routing Strategy.}

%


We consider our affinity-driven active selection routing as a hybrid of TCR \cite{clark2022unified,zhou2022mixture} and ECR. As the name suggested, TCR lets each token choose its \textit{top-scored} experts, and ECR lets each expert choose its \textit{top-scored} tokens. 
Specifically, we use the result of the expert-token affinity metrics as the affinity score between tokens and experts.
In conventional TCR routing strategy, the tokens are simply route to their Top-1 expert. In our hybird \textbf{TCR+ECR} routing strategy, experts also select tokens for processing from assigned tokens according to affinity scores:
\begin{equation}
	\small	
	\begin{aligned}
		&\left(\tilde{E}_{t1}, \dots, \tilde{E}_{t \ell} \right) = \text{Top-}\ell \left(\left\{ \delta_{t1}, \dots, \delta_{tn} \right\} \right), \\ 
		& ~ ~  \tilde{I}_{tk} \in [n], \forall t \in [s], k \in [\ell].
	\end{aligned}
\end{equation}
and then the expert to choose its Top-$\ell$ tokens where $\ell$ is determined by a threshold of the sum of affinity scores:
\begin{equation}
	\small	
	\begin{aligned}
		& \left(I_{1i}, \dots, I_{Ci} \right) = \text{Bottom-}C \left( \left\{t \in [s]: \exists j \in [\ell], \tilde{I}_{tj} = i \right\} \right), \\
		& ~ ~ I_{ki} \in [s] \cup \text{None}, \forall i \in [n], k \in [C].
	\end{aligned}
\end{equation}

Such bidirectional selection mechanism motivates each expert to receive a certain number of tokens with the highest affinity score to itself, thereby achieving a resonance effect. The resonance effect can help mitigate the homogenization in MoE.

\paragraph{Locality Loss.}

Feed-forward network (FFN) layers are commonly employed in expert networks, allowing each expert to learn independently as a separate neural network, thus preventing interference between samples. This mechanism leads to a severe load imbalance, as experts frequently selected in the early stages are more likely to be chosen in later stages. To mitigate this skewness in token allocation, the auxiliary loss \citep{shazeer2017outrageously} has been proposed. 
Building upon the auxiliary loss, our work introduces a loss bias term based on data locality, represented as $L_{\mathrm{loc}} = \mu \mathrm{KL}(D_{\mathrm{c}}||D_{\mathrm{l}}) = -\mu \int{D_{\mathrm{c}}(x)\ln[\frac{D_{\mathrm{l}}(x)}{D_{\mathrm{c}}(x)}]\mathrm{d}x}$, i.e., the Kullback-Leibler (KL) divergence of the current distribution $D_{\mathrm{c}}(x)$ and the fully localized distribution $D_{\mathrm{l}}(x)$. This loss term serves as a soft constraint, encouraging tokens to be sent to experts residing on the same node, thereby mitigating the substantial overhead incurred by partial inter-node communication.

\subsection{Training Strategy}

\paragraph{Token Distribution Dynamics under Expert Routing.}
The proposed hybrid TCR+ECR bidirectional routing mechanism operates through a two-stage process: tokens are initially assigned to experts based on the similarity between their feature fragments and the corresponding gating weights, followed by each expert selecting the Top-$\ell$ most relevant tokens through a scoring mechanism. This hybrid approach dynamically determines the number of tokens processed by each expert via an adaptive threshold, thereby ensuring the retention of class-discriminative tokens while optimizing computational efficiency.

\newtheorem{theorem}{Theorem}
\newtheorem{corollary}[theorem]{Corollary}
\newtheorem{assumption}[theorem]{Assumption}
\newtheorem{definition}[theorem]{Definition}
\newtheorem{lemma}[theorem]{Lemma}
\newtheorem{remark}[theorem]{Remark}

\subsubsection{Dynamic Lower Bound Module for Expert Capacity in ETR}\label{sec:theory}
To explain the motivation of our method, we show some theoretical insights in this section.

\begin{assumption}[data assumption]\label{ass:data}
	Each input $\vx \in \sR^{sd}$ with $s$ tokens is comprised of one class-discriminative pattern $\vo_1, \dots, \vo_n \in \sR^d$, with each decides the label in $[n]$, and $s-1$ class-irrelevant patterns $\vr \sim \gN$ for certain distribution $\gN$.
	For example, $\vx = (\vr_1, \vr_2, \vo_1, \vr_3, \dots, \vr_{s-1})$ has label $1$, where $\vr_i \stackrel{i.i.d.}{\sim} \gN, \forall i \in [s-1]$.
\end{assumption}

Based on Assumption \ref{ass:data}, \citet{chowdhury2023patch} demonstrated that the training of MoE go through two phases:

\textbf{Phase 1: Router training} This process ensures that each expert only receives the class-discriminative tokens related to the specific class.

\textbf{Phase 2: Expert training} This process is designed to establish each expert's ability to handle and solve problems. 

To quantitatively measure the difference between TCR and ECR, we define \textbf{training success rate} of input motivated by the training process of MoE. 

\begin{definition}[training success rate]
	We say the input $\vx \in \sR^{sd}$ with $s$ tokens succeed in training if
	the class-discriminative pattern in $\vx$, e.g., $\vo_i$ is correctly dispatched to $i$-th expert.
	We further define \textbf{training success rate} as the probability that the input succeed in training.
\end{definition}

Furthermore, to show the quantitative comparison of TCR and ECR in training success rate, we need following asssumptions and notations of token patterns.

\begin{assumption}[class-discriminative]\label{ass:dis}
	We assume the location and feature of class-discriminative pattern is uniformly distribute in $[s]$ and $[n]$, i.e.,
	\begin{equation}
		\small
		i \sim \mathrm{Unif}([s]), \vx_i \sim \mathrm{Unif}\left(\{\vo_1, \dots, \vo_n\}\right).
	\end{equation}
	We also assume that $\forall i \in [n], \vo_i$ should be sent to the $i$-th expert, and define the true positive probability in token choice setting is no worse than the uniform dispatch as below
	\begin{equation}
		\small	
		\sP(\delta_{\vo_i, i} \geq \delta_{\vx_j, i}, \forall j \in [s]) = p_i \geq 1/n, \forall i \in [n]. 
	\end{equation}
\end{assumption}

\begin{assumption}[class-irrelevant]\label{ass:irr}
	The distribution of class-irrelevant patterns is isotropy, i.e., 
	\begin{equation}\label{eq:irr}
		\small	\sP(\vr \sim \gN, \delta_{\vr, i} \geq \delta_{\vx_j, i}, \forall j \in [s]) = 1/n, \forall i \in [n].
	\end{equation}
	And we define the false positive probability in expert choice setting as
	\begin{equation}
		\small	\sP(\vr \sim \gN, \delta_{\vr, i} \geq \delta_{\vo_i, i}) = q_i, \forall i \in [n],
	\end{equation}
	which measures the possibility that expert $i$ chooses the wrong token $\vr$ instead of the correct token $\vo_i$.
\end{assumption} 

\begin{theorem}\label{thm:main}
	Under Assumptions \ref{ass:dis} and \ref{ass:irr}, the training success rate of TCR in each sample $\vx$ is
	\begin{equation}\label{eq:tcr}
		\small	\sP(\text{TCR succeed}) = \Theta\big(C\sum_{i=1}^n p_i/s \big),
	\end{equation}
	and the training success rate of ECR is $ \forall i \in [n]$,
	\begin{equation}\label{eq:ecr}
		\small	\sP(\text{ECR succeed}) \begin{cases}
			\leq \frac{1}{n}\sum_{i=1}^n e^{-\frac{(s-1)q_i}{8}}, &C \leq (s - 1) q_i/2, \\
			\geq 1 - e^{-3C/16}, &C \geq 2sq_i.
		\end{cases}
	\end{equation}
\end{theorem}

\begin{corollary}
	In practice, For constant number of experts \cite{jiang2024mixtral}, i.e., $n=\Theta(1)$, and $C < s$ to save computation cost.
	We have the following lower bound for capacity $C$ to ensure high training success rate:
	\begin{enumerate}
		\item Suppose $q_i=\Theta(1)$. Then TCR is much better than ECR, and we only need $C=\Theta(s)$.
		\item Suppose $\forall i \in [n], sq_i\leq C^*$ for some $C^* > 0$. Then ECR is much better than TCR, and we only need $C \geq 2C^*$.
	\end{enumerate} 
\end{corollary}

\begin{remark}\label{remark:com}
	Under the assumptions of orthogonal gating weights and uniform data distribution, we establish that the lower bound of expert capacity is given by $C_{\mathrm{min}} = \frac{1}{n}\exp\{d\delta_{\mathrm{max}}^2/(2-\delta_{\mathrm{max}}^2)\}$, where the capacity is intrinsically linked to the angular relationship between gating weights and tokens.
	
	Building on Theorem \ref{thm:main} and the evolution of feature distributions during training, we demonstrate the optimality of transitioning from TCR to ECR. In early training stages, when class-irrelevant tokens exhibit near-isotropic distribution with $q_i = \Theta(1)$, TCR achieves superior training success rates of $C/s$ compared to ECR's exponentially decaying rate of $e^{-s}$, necessitating a larger expert capacity of $C = \Theta(s)$. As training progresses and experts develop discriminative capabilities, the distribution shifts such that $q_i \ll 1$ or $sq_i \leq C^*$ for some constant $C^* > 0$, at which point ECR approaches unit success rate while TCR remains bounded by $C/s$, enabling efficient operation with reduced capacity $C = \Theta(1)$ when $C \geq 2C^*$.
\end{remark}

\subsection{Communication Optimization}

To mitigate the serial execution bottlenecks inherent in LLM training, we implement Communication Over Computation (CoC) optimization, which transforms sequentially dependent matrix multiplication and collective communication operations in parallel linear layers into unified, fine-grained kernels. By leveraging MTE's remote memory access capabilities, CoC fuses computation and communication primitives, enabling pipeline-parallel execution that overlaps previously serialized operations.

\section{Experiments}
\label{experiment}

\subsection{Experimental Setup}

We implement our approach using Mixtral 8×7B, a 46.7B-parameter model with Group Query Attention (GQA) and 32 sparse MoE blocks. Each block contains 8 experts, with tokens routed to their top-2 selections. To accommodate long-context applications, we extend the sequence length to 32,768 tokens. Our experiments span three cluster scales with tailored parallelization strategies: 32 NPUs (TP=4, PP=4, DP=2, EP=2), 64 NPUs (TP=8, PP=4, DP=2, EP=2), and 256 NPUs (TP=8, PP=8, DP=4, EP=2), maintaining a global batch size of 128 throughout. Additional experimental details are provided in the Appendix.

\subsection{Efficiency Promotion and Memory Footprint Reduction}

We consistently employ Top-1 routing to align implementation with our theoretical framework. The baseline model uses constrained expert capacity rather than groupedGEMM, preventing token dropout with a capacity factor of 1.1. LocMoE incorporates distributional uniformity and estimates expert capacity via a theoretically-derived lower bound formula from the initial batch, maintaining this value throughout training. Our approach ("LocMoE+" in figures) constrains score sum ranges, processes hidden states, and dynamically calculates expert capacity. The subsequent analysis examines training efficiency, convergence, and memory utilization across multiple Ascend cluster configurations.

\begin{figure}[H]
	\centering
	\includegraphics[width=0.48\textwidth]{figures/time_v2.png}
	\caption{The time consumption during training iterations with different schemes and cluster sizes.}
	\label{time}
\end{figure}

Figure \ref{time} presents the training overhead analysis across the initial 1000 training iterations. To ensure measurement stability and exclude initialization artifacts, we commence time profiling from the fifth iteration onward. The baseline model exhibits consistent temporal performance throughout the evaluation period. In contrast, LocMoE demonstrates a marginal decrease in execution time as training progresses, with this trend being particularly pronounced in the 32N and 64N configurations. This observation corroborates our hypothesis that locality-aware optimization achieves optimal efficiency when the number of experts meets or exceeds the number of computational nodes. Our proposed method introduces a modest computational overhead relative to LocMoE, attributable to the token rearrangement mechanism. However, this overhead diminishes progressively as token representations converge during training. Specifically, the convergence of token features leads to a reduction in the number of tokens requiring rearrangement, resulting in stabilized computational costs in later training phases. Empirically, our approach achieves a reduction in total training time ranging from 2.9\% to 13.3\% compared to LocMoE, and from 5.4\% to 46.6\% relative to the baseline configuration.

\begin{figure}[H]
	\centering
	\includegraphics[width=0.40\textwidth]{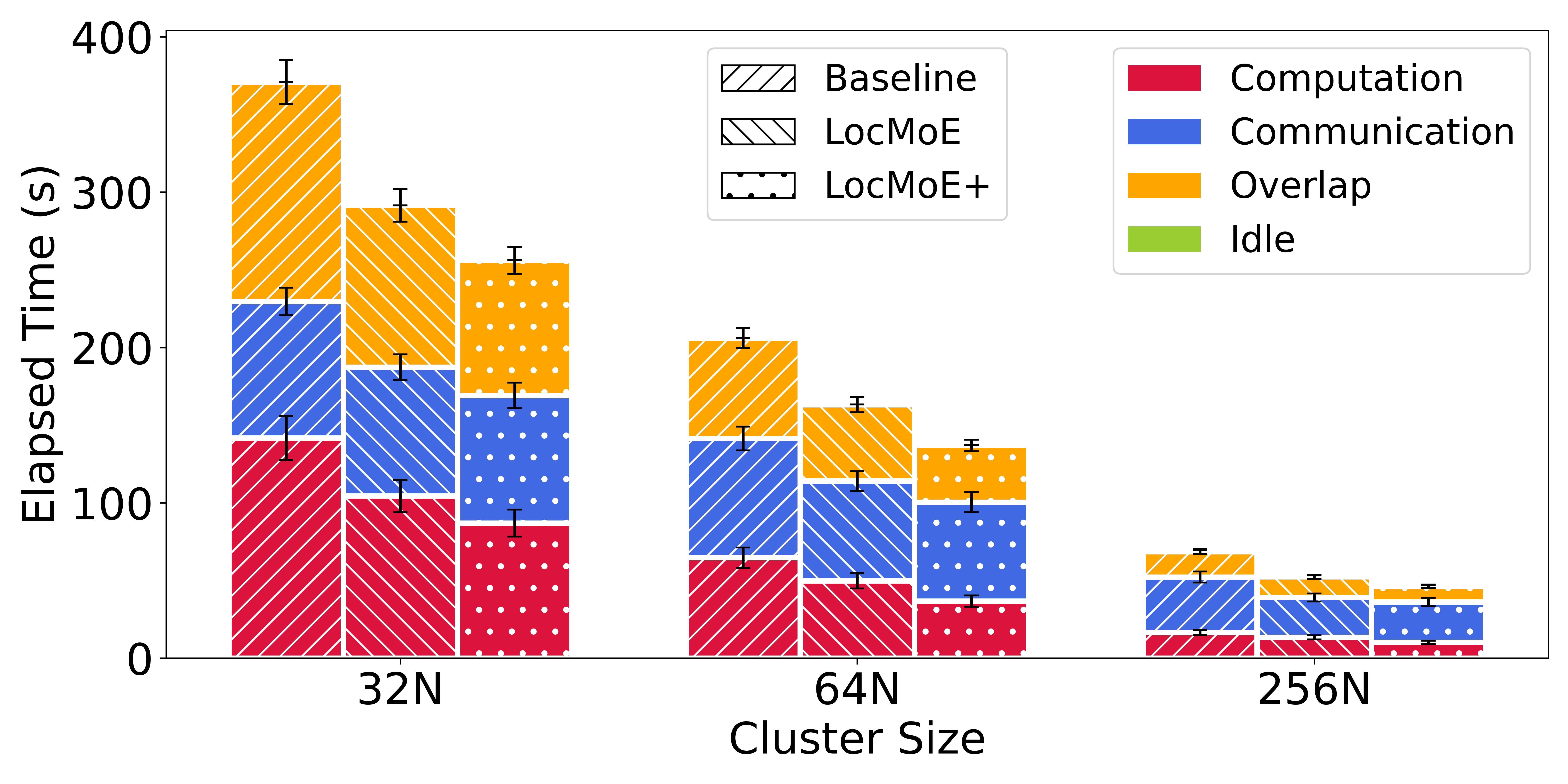}
	\caption{The average composition of computation, communication, overlap, and idle with different schemes and cluster sizes.}
	\label{time_ratio}
\end{figure}

Figure \ref{time_ratio} depicts the temporal distribution of computational phases across training. We sample performance metrics at ten equidistant intervals throughout training, capturing computation, communication, overlap, and idle time. While profiling introduces minor overhead, this methodology provides robust insights into system behavior.

Both LocMoE and our proposed method demonstrate reduced latency across all components, with computational overhead exhibiting more substantial improvements than communication costs. This efficiency gain follows a clear pattern: as cluster size expands, the computation-communication overlap ratio decreases, accompanied by diminishing returns in computational speedup. This trend reflects the inherent scalability challenges in distributed MoE architectures.

\begin{figure}[H]
	\centering
	\includegraphics[width=0.40\textwidth]{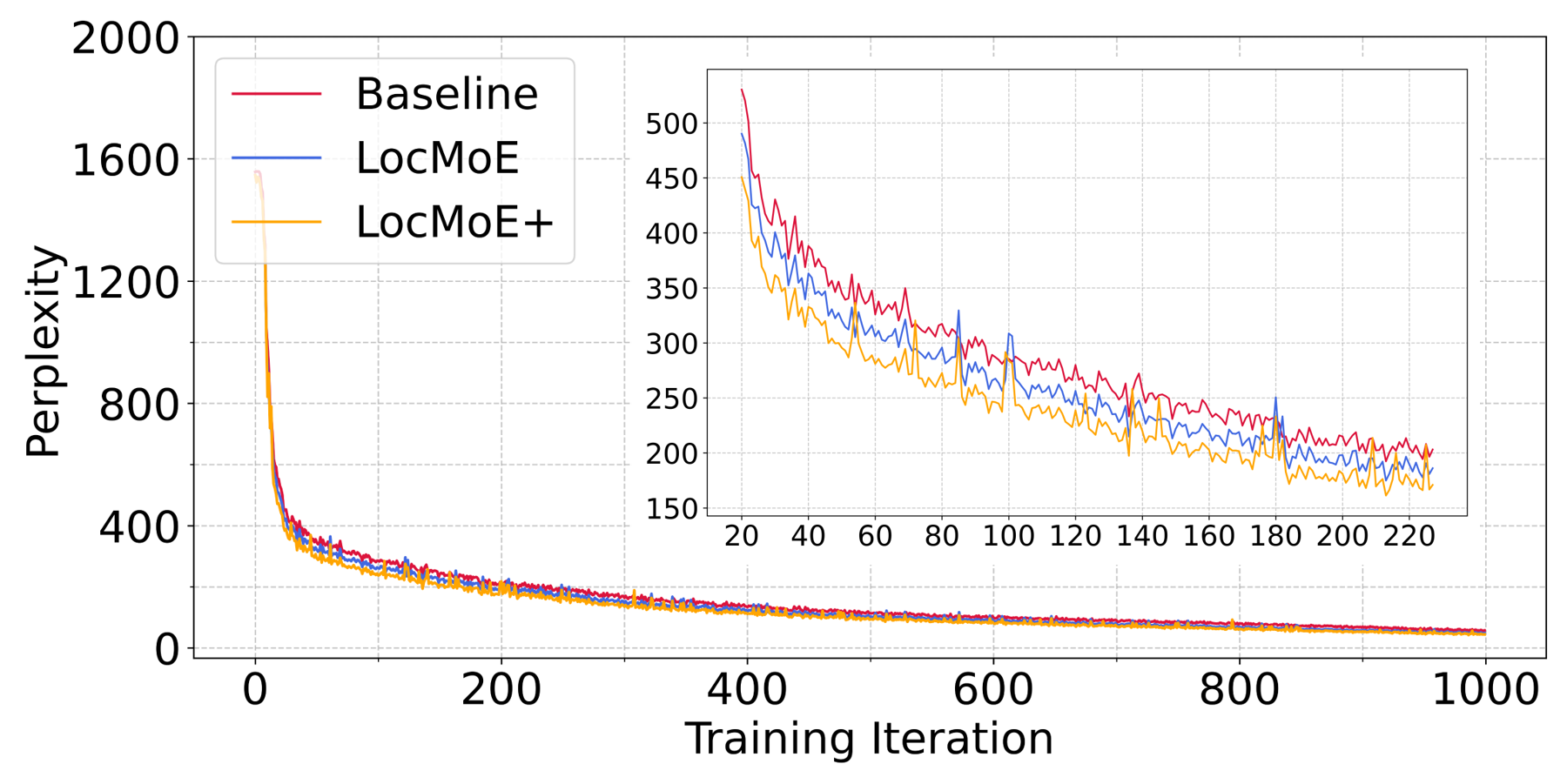}
	\caption{The perplexity during training iterations with different schemes.}
	\label{ppl}
\end{figure}

Figure \ref{ppl} validates that these efficiency improvements preserve model quality. All methods exhibit comparable convergence trajectories, confirming that our optimization strategy maintains training stability while delivering performance gains. The perplexity curves demonstrate that accelerated training does not compromise the fundamental learning dynamics of the model.

Figure \ref{op} presents the operator-level computational profiling across different hardware components. The system leverages AI CORE for matrix multiplication and convolution operations, AI VECTOR CORE for parallelized vector computations, MIX AIC for heterogeneous operator fusion, and AI CPU for specialized AI instruction execution. Our token selection strategy yields substantial performance gains: the FFN MatMul operator achieves a 17× speedup compared to the baseline and 2.6× improvement over LocMoE. This optimization translates to a 2.8× reduction in cumulative MatMul execution time and a 2.6× decrease in Cube computational load. The rearrangement-associated operators (TopK and IndexPutV2) exhibit marginal overhead increases, representing an acceptable trade-off for the significant computational savings achieved through selective token processing.

\begin{figure}[H]
	\centering
	\includegraphics[width=0.5\textwidth]{figures/op_ratio_v2.png}
	\caption{The distribution of time consumption for operators.}
	\label{op}
\end{figure}

Figure \ref{footprint} analyzes memory consumption patterns during stable training phases, based on 100,000 memory profiling samples per device. Our approach demonstrates substantial memory efficiency gains, achieving 4.57-16.27\% reduction compared to the baseline and 2.86-10.5\% reduction compared to LocMoE. The memory optimization exhibits scale-dependent characteristics: larger clusters show reduced computational overhead proportions and correspondingly narrower memory usage differentials. Furthermore, our method effectively eliminates transient memory spikes and reduces short-term memory fluctuations, contributing to more predictable and stable resource utilization throughout training.

\begin{figure}[H]
	\centering
	\includegraphics[width=0.48\textwidth]{figures/Footprint_v2.png}
	\caption{
		print recorded in one acquisition cycle with different schemes and cluster sizes.}
	\label{footprint}
\end{figure}

\subsection{Expert Homogenization and Load Distribution Analysis}

The Calinski-Harabasz (CH) index \citep{lima2020genetic} measurements reveal that bidirectional affinity selection significantly enhances token clustering quality in MoE architectures:

\begin{equation}
	CH = \frac{\sum_{i=1}^{k} n_i ||c_i - c||^2}{\sum_{i=1}^{k} \sum_{x \in C_i} ||x - c_i||^2}
\end{equation}

\begin{figure}[H]
	\centering
	\includegraphics[width=0.45\textwidth]{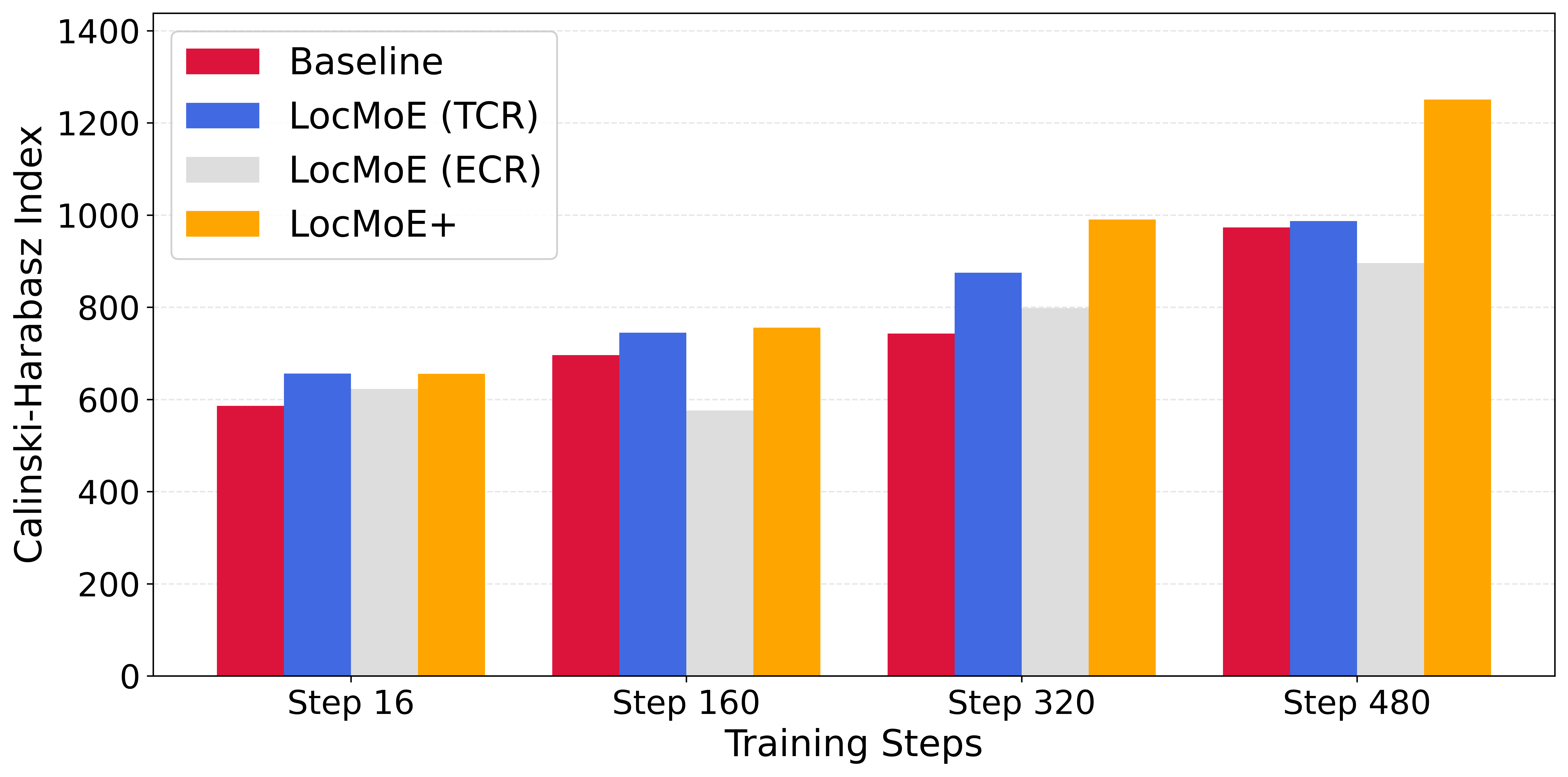}
	\caption{The Calinski-Harabasz index across training steps.}
	\label{ch_index}
\end{figure}

While baseline and single-mechanism approaches achieve comparable improvements, the integrated LocMoE+ method demonstrates superior performance, as shown in Figure \ref{ch_index}. Combining token-to-expert and expert-to-token selection mechanisms creates synergistic effects that accelerate natural clustering tendencies during training. Our bidirectional approach establishes a positive feedback loop between token routing and expert affinity, fundamentally enhancing expert utilization efficiency through improved feature organization and specialization.

Figure \ref{loss_distribution} reveals fundamental differences in how various loss functions affect token distribution across experts. The baseline approach suffers from severe load imbalance, with certain experts becoming overloaded while others remain idle. The auxiliary loss method provides marginal improvements through regularization, yet distribution remains significantly skewed. The locality loss demonstrates transformative effects by incorporating architectural topology into the optimization objective, achieving balanced token allocation across all experts through KL divergence constraints that simultaneously minimize inter-node communication and prevent expert collapse. 

\begin{figure}[H]
	\centering
	\includegraphics[width=0.48\textwidth]{figures/moe_expert_token_distribution.png}
	\caption{The distribution of tokens assigned to experts with different loss function.}
	\label{loss_distribution}
\end{figure}

Figure \ref{cdf_ecdf} presents the cumulative distribution function (CDF) and empirical cumulative distribution function (ECDF) analysis across these routing methods throughout the training progression. The locality loss approach presents distinctly optimal characteristics across both CDF and ECDF measurements, maintaining consistently high performance levels throughout the training process with remarkable stability during later training phases. The sustained performance across different probability and load thresholds indicates that incorporating expert-token affinity into the routing objective creates robust optimization dynamics that preserve both routing quality and load distribution efficiency. These findings underscore the effectiveness of ETR in addressing the inherent challenges of expert-token assignment optimization, providing a principled foundation for scalable sparse model.

\begin{figure}[H]
	\centering
	\includegraphics[width=0.48\textwidth]{figures/cdf_ecdf_comparison.png}
	\caption{The CDF and ECDF of different schemes.}
	\label{cdf_ecdf}
\end{figure}

\subsection{The Performance of Downstream Tasks}
The GDAD benchmark, comprising three distinct evaluation tasks, serves as the primary assessment framework for domain task capabilities, while TeleQnA provides specialized validation for telecommunications knowledge applications. Our integrated LocMoE+ approach consistently outperforms both baseline and single-mechanism variants across all evaluation metrics (see Table \ref{tab:results}), achieving notable gains through bidirectional selection mechanisms. The consistent improvements observed across both general domain tasks and specialized telecommunications applications confirm that the synergistic combination of bidirectional selection mechanisms creates substantial performance advantages while validating the robustness and generalizability of our proposed architecture modifications.

\begin{table}[H]
	\caption{Domain performance promotion obtained by our approach on different datasets.}
	\label{tab:results}
	\centering
	\setlength{\tabcolsep}{2pt}
	\begin{tabular}{lrrrrrr}
		\toprule
		& \multicolumn{4}{c}{GDAD} & \multicolumn{1}{c}{} \\
		\cmidrule{2-4}
		& \multicolumn{1}{c}{GDAD-1} & \multicolumn{1}{c}{GDAD-2} & \multicolumn{1}{c}{GDAD-3} & \multicolumn{1}{c}{Avg} & \multicolumn{1}{c}{TeleQnA} \\
		\midrule
		Baseline         & 47.8    & 43.0   & 65.4   & 52.8  & 62.1 \\
		LocMoE (TCR)          & 55.5    & 47.6   & 71.1   & 59.0  & 67.6  \\
		LocMoE (ECR)          & 45.8    & 45.6   & 62.8   & 56.3  & 61.8  \\
		\textbf{LocMoE+} & \textbf{57.4}    & \textbf{49.9}   & \textbf{74.5}   & \textbf{61.5}  & \textbf{68.8}  \\
		\bottomrule
	\end{tabular}
\end{table}

Table \ref{tab:performance_comparison} presents general performance evaluation results across three widely-recognized benchmarks—MMLU \citep{hendryckstest2021} for comprehensive knowledge assessment, GPQA \citep{rein2023gpqa} for advanced reasoning capabilities, and HumanEval \citep{chen2021evaluating} for code generation proficiency—revealing distinct performance characteristics of our proposed methods. The results demonstrate that our bidirectional LocMoE+ approach achieves superior performance in reasoning and coding tasks while maintaining competitive general knowledge capabilities, with individual constraint routing mechanisms exhibiting complementary strengths across different evaluation dimensions. While the baseline maintains slight advantage in MMLU, the integrated LocMoE+ approach demonstrates that bidirectional selection mechanisms create meaningful improvements in task-specific capabilities without substantial degradation in general knowledge retention, suggesting that our architectural modifications enhance model specialization for complex reasoning and generation tasks while preserving foundational knowledge capabilities.

\begin{table}[H]
	\centering
	\caption{General performance comparison of different MoE methods}
	\label{tab:performance_comparison}
	\begin{tabular}{lccc}
		\toprule
		Method & MMLU & GPQA & HumanEval \\
		\midrule
		Baseline & \textbf{71.8} & 29.2 & 40.2 \\
		LocMoE (TCR) & 68.4 & 30.3 & 52.8 \\
		LocMoE (ECR) & 45.8 & 32.5 & 57.6 \\
		LocMoE+ & 70.4 & \textbf{33.5} & \textbf{67.8} \\
		\bottomrule
	\end{tabular}
\end{table}

To enhance conversational capabilities and downstream task adaptability, we conducted supervised fine-tuning on the pre-trained models. As shown in Figure \ref{radar}, our approach demonstrates substantial improvements across multiple evaluation dimensions within the General and Domain-specific Assessment Dataset (GDAD). The method achieves an average improvement of approximately 20.1\% across 16 sub-capabilities of Domain Task Capability compared to the baseline, with particularly notable gains in rewriting and summary capabilities. In the Domain Competency Exam assessments, our approach shows an average improvement of 16\% relative to the baseline, with IP Training in digital communications demonstrating the most significant advancement. Among the 18 sub-capabilities of General Ability, the method exhibits an improvement of about 13.9\% relative to the baseline, with planning capabilities showing the highest enhancement at 26.8\%.

\begin{figure}[H]
	\centering
	\includegraphics[width=0.5\textwidth]{figures/radar_v2.png}
	\caption{The performance on three categories of GDAD.}
	\label{radar}
\end{figure}

\section{Conclusion}

In this paper, we propose ETR, a fundamentally new approach to MoE routing that solves the longstanding trade-off between computational efficiency and model performance through theoretically-grounded bidirectional selection mechanisms. By dynamically coordinating token-choice and expert-choice routing based on training progress, ETR achieves simultaneous improvements in both training efficiency and downstream task quality. The substantial performance gains demonstrated across diverse benchmarks, combined with significant reductions in computational overhead, establish ETR as a critical advancement for practical deployment of large-scale sparse models. Our theoretical contributions provide new insights into expert-token dynamics that extend beyond incremental optimizations, opening pathways for next-generation MoE architectures.

\bibliography{custom}

\clearpage
\section{Appendix}
\label{app:mis-proof}
\section{Missing Proof}
\subsection{Auxiurary Results}
\begin{lemma} \label{lemma:aux}
	Let  $\mX_1, \dots, \mX_n$ be $n$ independent random variables with
	\begin{equation} 
		\small \sP(\mX_i=1)=p_i, \sP(\mX_i=0) = 1 - p_i.
	\end{equation}
	We consider the sum $\mX = \sum_{i=1}^n X_i$, with expectation $\E(X) = \sum_{i=1}^n p_i$. Then we have
	\begin{equation}
		\small
		\begin{aligned}
			& \mathrm{(Lower \ tail)  } \quad \sP(\mX \leq \E\mX -\lambda) \leq e^{-\frac{\lambda^2}{2 \E\mX }}, \\
			& \mathrm{(Upper \ tail)  } \quad \sP(\mX \geq \E\mX +\lambda) \leq e^{-\frac{\lambda^2}{ 2(\E \mX + \lambda/3)}}.
		\end{aligned}
	\end{equation}
\end{lemma}

\subsection{Proof of Theorem 5}
\begin{proof}
	1) For the TCR, denote 
	\begin{equation}
		\small s_i = \left|\left\{t < k: \vx_t \text{ sent to expert } i, \vx_k = \vo_i \right\} \right|, \forall i \in [n]
	\end{equation}
	as the top class-irrelevant token number candidated to the $i$-th expert before the valid token.
	Then by Assumption \ref{ass:irr}, each class-irrelevant token uniformly gives to any expert, leading to $s_i|(\vx_k=\vo_i) \sim \gB(k-1, 1/n)$ (Binomial distribution), i.e., $\forall t \in [k-1]$,
	\begin{equation}
		\small \sP(s_i = t| \vx_k = \vo_i) = \begin{pmatrix}
			k-1 \\ t
		\end{pmatrix} \cdot \left(\frac{1}{n}\right)^{t} \left(1-\frac{1}{n}\right)^{k-1-t}. 
	\end{equation}
	Then we could derive that
	\begin{equation*}
		\small
		\begin{aligned}
			&\sP(\vx \text{ succeed in training}) \nonumber  \\
			&= \sum_{i=1}^n \sP(\vo_i \text{ sent to expert } i| \vo_i \text{ is in } \vx) \cdot \sP(\vo_i \text{ is in } \vx) \nonumber \\
			&= \frac{1}{ns} \sum_{i=1}^n \sum_{k=1}^s p_i \sP(s_i < C | \vx_k=\vo_i) \nonumber \\
			&= \frac{1}{ns} \sum_{i=1}^n p_i \left(C + \sum_{k=C + 1}^s \sP(s_i < C | \vx_k=\vo_i) \right).  \label{eq:tmp-1}
		\end{aligned}
	\end{equation*}
	Note that $\E s_i = (k-1)/n$. 
	When $k \geq 2nC$, by lower tail bound in Lemma \ref{lemma:aux}, we get
	\begin{equation}
		\small \sP(s_i < C | \vx_k=\vo_i) \leq e^{-\frac{(k-1-n(C-1))^2}{2(k-1)n}} \leq e^{-\frac{k-1}{8n}}.
	\end{equation}
	Hence, we get the upper bound that
	\begin{equation*}
		\small
		\begin{aligned}
			&\sP(\vx \text{ succeed in training}) \\
			&\stackrel{\eqref{eq:tmp-1}}{\leq} 
			\frac{1}{ns} \sum_{i=1}^n \sum_{k=1}^s p_i \sP(s_i < C | \vx_k=\vo_i) \\
			&= \frac{1}{ns} \sum_{i=1}^n p_i \left(2nC + \sum_{k=2nC+1}^s \sP(s_i < C | \vx_k=\vo_i) \right) \\
			&\leq \frac{1}{ns} \sum_{i=1}^n p_i \left(2nC + \sum_{k=2nC}^{s-1} e^{-\frac{k}{8n}} \right) \\
			&\leq \frac{1}{ns} \sum_{i=1}^n p_i \left(2nC + \frac{e^{-\frac{C}{4}}}{1-e^{-\frac{1}{8n}}} \right) \\
			&\stackrel{(i)}{\leq} \frac{1}{ns} \sum_{i=1}^n p_i \left(2nC + (8n+1) e^{-\frac{C}{4}} \right) \leq \frac{10 C\sum_{i=1}^n p_i }{s},
		\end{aligned}
	\end{equation*}
	where $(i)$ uses the inequality that $e^{-t} \leq 1/(1+t), \forall t \geq 0$.
	
	Moreover, for $1 + \frac{nC}{4} \leq k \leq 1 + \frac{nC}{2}$, i.e., $2(k-1) \leq nC \leq 4(k-1)$, by upper tail bound in Lemma \ref{lemma:aux}, we get
	\begin{equation*}
		\small
		\begin{aligned}
			\sP(s_i < C | \vx_k=\vo_i) &= 1 - \sP(s_i \geq C | \vx_k=\vo_i) \\
			&\geq 1 - e^{-\frac{3(nC - k + 1)^2}{2n[2(k-1)+nC]}} \geq 1 - e^{-\frac{k-1}{4n}}.
		\end{aligned}
	\end{equation*}	
	Hence, we get the lower bound that
	\begin{equation*}
		\small
		\begin{aligned}
			&\sP(\vx \text{ succeed in training}) \\
			& \stackrel{\eqref{eq:tmp-1}}{\geq} 
			\frac{1}{ns} \sum_{i=1}^n \sum_{k=1}^s p_i \sP(s_i < C | \vx_k=\vo_i) \\
			&= \frac{1}{ns} \sum_{i=1}^n p_i \left(\sum_{k=\lceil 1+nC/4 \rceil }^{\lfloor 1+nC/2 \rfloor } \sP(s_i < C | \vx_k=\vo_i) \right) \\
			&\geq \frac{1}{ns} \sum_{i=1}^n p_i \left(\frac{nC}{4} - 1 - \sum_{k=\lceil 1+nC/4 \rceil }^{\lfloor 1+nC/2 \rfloor }  e^{-\frac{k-1}{4n}} \right) \\
			& \geq \frac{1}{ns} \sum_{i=1}^n p_i \left(\frac{nC}{4} - 1 - \frac{e^{-\frac{C}{16}}}{1-e^{-\frac{1}{4n}}} \right) \\
			&\stackrel{(i)}{\geq} \frac{1}{ns} \sum_{i=1}^n p_i \left(\frac{nC}{4} - 2 - (4n+1) e^{-\frac{C}{16}} \right) \geq \frac{C\sum_{i=1}^n p_i }{5s},
		\end{aligned}
	\end{equation*}
	where $(i)$ uses the inequality that $e^{-t} \leq 1/(1+t), \forall t \geq 0$, and the final inequality needs $C \geq 48$, which can be satisified in common experiments.
	Combining the upper and lower bounds, we obtain the desired result.
	
	2) For the ECR, denote $s_i$ as the class-irrelevant token number with the score larger than $\vo_i$ for $i$-th expert. 
	By Assumption 4, we derive that $s_i \sim\gB(s-1, q_i), \forall i \in [n]$.
	\begin{equation*}
		\small
		\begin{aligned}
			&\sP(\vx \text{ succeed in training}) \\
			&= \sum_{i=1}^n \sP(\text{expert } i \text{ choose } \vo_i| \vo_i \text{ is in } \vx) \sP(\vo_i \text{ is in } \vx) \\
			&= \frac{1}{n} \sum_{i=1}^n \sP(s_i \leq C - 1, s_i \sim \gB(s-1, q_i))
		\end{aligned}
	\end{equation*}	
	If $C - 1 \leq (s - 1) q_i / 2$, by lower tail bound in Lemma \ref{lemma:aux} with $\lambda = (s-1)q_i-(C-1) <\E s_i $, we obtain that
	\begin{equation}
		\small \sP(s_i \leq C-1) \leq e^{-\frac{(s-1)q_i}{2}\left(1-\frac{C-1}{(s-1)q_i}\right)^2} \leq e^{-\frac{(s-1)q_i}{8}}.
	\end{equation}	
	If $C \geq 2(s-1) q_i $, by upper tail bound in Lemma \ref{lemma:aux} with $\lambda = C-(s-1)q_i > 0$, we obtain that
	\begin{equation*}
		\small
		\begin{aligned}
			\sP(s_i \leq C - 1) &= 1 - \sP(s_i \geq C) \\
			&\geq 1 - e^{-\frac{[C-(s-1)q_i]^2}{2(C+2(s-1)q_i)/3}} \geq 1 - e^{-\frac{3C}{16}}.
		\end{aligned}
	\end{equation*}	
	Hence, we conclude Eq.(10).
\end{proof}

\section{Token Feature Distribution}\label{app:token}
\begin{figure}[t]
	\centering
	\includegraphics[width=0.5\textwidth]{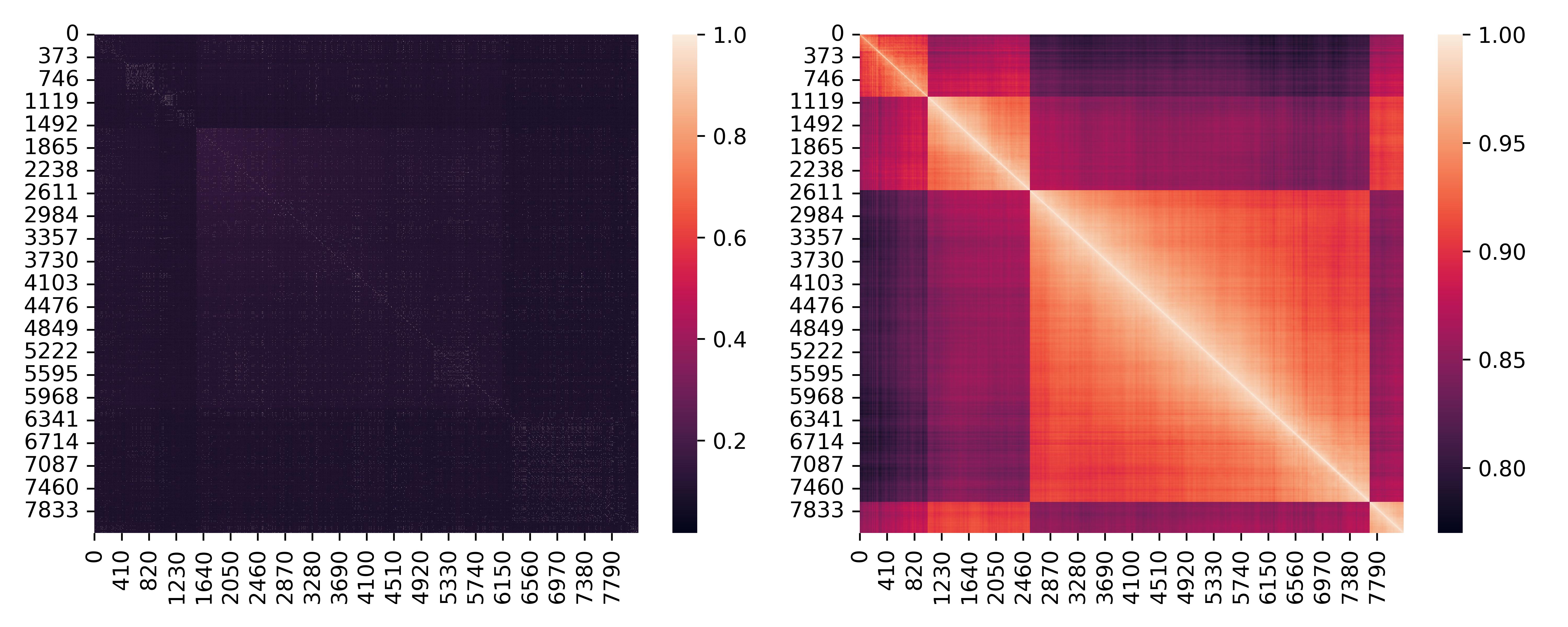}
	\caption{The correlation matrix of one training sample feature before (left) and after (right) training.}
	\label{fig:token-dis}
\end{figure}

We also validate the feature distribution before and after MoE training shown in Figure \ref{fig:token-dis}. 
We can see before training, all 8192 tokens in one training sample are nearly orthogonal with correlation coefficient near zero, which verifies the isotropy distribution assumption in the first bullet of Remark 7.
After training, the token features are nearly aligned with correlation coefficien large than $0.8$. 
We can also observe that neighbouring tokens share similar features, and clear block feature behavior, meaning that the token features are relatively separated and the number of tokens in each cluster is bounded, which somehow matches the distribution assumption in the second bullet of Remark 7.

\section{Experimental Setup}\label{setup}
\subsection{Datasets for Pre-training and Fine-Tuning}

The dataset used in this paper is a self-constructed dataset that integrates knowledge from multiple domains, including wireless, data communication, and cloud-core technologies. It comprises Chinese, English, and bilingual corpora. The corpora are parsed from various internal technical documents, such as iCase, blogs, Wiki, and feature documents. Taking iCase as an example, iCase is a case record of problem localization and handling processes, containing code, instructions, and corresponding logs. In addition, the above-mentioned domain-specific knowledge corpora are mixed with general corpora in a ratio of 1:5. The general corpora are collected from hundreds of websites, including online novels, cooking guides, movie reviews, and more. After cleaning, deduplication, and review operations, the dataset is thoroughly shuffled. A total of 4.19 billion tokens is sampled as the experimental pre-training dataset. To evaluate downstream tasks, this paper also adopt hybrid sft data items to fine-tune the pre-trained model. The dataset comprises 762,321 general question-answer pairs and 11,048 domain-specific question-answer pairs, with a general-to-domain ratio of 68:1. The general characteristics encompass multi-tasking, mathematical ability, coding ability, logical reasoning, multi-turn dialogue, knowledge reasoning, language understanding, text generation, multi-tasking, FunctionCall, CoT, MRC summarization, refusal to answer, Chinese, and English. The domain-specific characteristics include domain knowledge understanding, RAG, FunctionCall, information extraction, multi-turn dialogue, reading comprehension, paraphrasing, and intent recognition.

The pre-training data comprises 300B tokens in total, with 150B tokens from the ICT domain and 150B tokens sampled from general data. The sampling ratios are shown in Table \ref{data_mixing}. For SFT data, we employed a two-stage training: the first stage primarily enhances the model's logical reasoning capabilities such as multi-task capability, mathematics, puzzle-solving, complex logic, etc. The total scale of samples is approximately two million, while the second stage focuses on improving instruction-following abilities, tool function call, sentiment, security, etc. The total scale of samples in stage2 is about three million.

\begin{table*}[htbp]
	\centering
	\caption{Data sources and sampling ratios of general pre-training data.}
	\begin{tabular}{llll}
		\toprule
		\textbf{Primary Category} & \textbf{Secondary Category} & \textbf{Tertiary Source} & \textbf{Sampling ratio} \\
		\midrule
		\multirow{11}{*}{General English} & \multirow{2}{*}{Webpages} & Reasoning\_steplist & 25\% \\
		& & Model\_rewrite & 100\% \\
		& \multirow{6}{*}{Books \& Papers} & book3 & 25\% \\
		& & bookcorpus & 100\% \\
		& & all\_libgen\_books & 20\% \\
		& & all\_libgen\_scihub & 10\% \\
		& & RedPajama\_arxiv & 25\% \\
		& & arxiv\_latex2Markdown\_cleaned & 25\% \\
		& \multirow{3}{*}{WebText} & wiki & 100\% \\
		& & stackexchange\_cleaned & 20\% \\
		& & cosmopedia\_v2 & 15\% \\
		\multirow{9}{*}{General Chinese} & Webpages & aigc\_dataset & 15\% \\
		& \multirow{4}{*}{Book} & all\_book\_deduped & 10\% \\
		& & zh\_book\_CommonData & 10\% \\
		& & zh\_general\_STEM & 80\% \\
		& & all\_zhiwang & 20\% \\
		& \multirow{4}{*}{WebText} & baike\_MBAzhiku\_sougou\_ye\_zhiarge & 50\% \\
		& & baike\_sougou\_baidu\_kuaidong & 50\% \\
		& & wiki & 10\% \\
		& & zhihu\_caigou\_merged\_cleaned & 10\% \\
		\multirow{5}{*}{High-density Knowledge} & Q\&A & quiz\_data & 100\% \\
		& \multirow{4}{*}{Collection} & density\_knowledge & 100\% \\
		& & collection\_updated & 100\% \\
		& & english\_question\_and\_answer & 100\% \\
		& & annealing & 100\% \\
		\multirow{3}{*}{Code} & Code & code\_python\_edu\_high\_quality & 30\% \\
		& \multirow{2}{*}{Forum} & CSDN & 20\% \\
		& & Ultra\_textbooks & 100\% \\
		\bottomrule
	\end{tabular}
	\label{data_mixing}
\end{table*}

\subsection{Experimental Environment}

The experiments are conducted on a cluster composed of Ascend 910B3 NPUs, divided into three groups: 32 NPUs (hereinafter referred to as 32N, and so on), 64N, and 256N. The 910B3 series NPU contains 20 AI cores with a main frequency of 1.8GHz and a theoretical computing power of 313T under fp16 precision. The physical High Bandwidth Memory (HBM) of the 910B3 NPU is 64G, with an HBM frequency of 1.6GHz and an HBM bandwidth of 1.6T. Every 8 NPUs are mounted on the same Atlas 800T A2 server, which internally adopts a fullmesh networking scheme, meaning that any two NPUs are interconnected.

\subsection{Evaluation Metrics and Datasets}

To evaluate model performance, this paper designs a comprehensive metric called the General and Domain-specific Assessment Dataset (GDAD), which consists of three evaluation systems: domain task capability, domain capability certification exam, and general capability. Among them, the domain task capability includes a total of 16 categories and 2,657 questions, such as domain logical reasoning; the domain capability certification exam includes a total of 13 categories and 13,968 questions, such as data communication; and the general capability includes a total of 18 categories and 1,435 questions, such as programming ability. The questions include objective and subjective questions in Chinese, English, and bilingual formats. For subjective questions, the cosine similarity between the model output and the standard answer is used as the score. In addition, this paper also employs GPQA \citep{rein2023gpqa} and TeleQnA \citep{maatouk2023teleqna} to evaluate the model's Chinese language capability.

\end{document}